# Spatial Relationship Based Features for Indian Sign Language Recognition

Chethana Kumara B M, Nagendraswamy H S and Lekha Chinmayi R

*Abstract*— In this paper, the task of recognizing signs made by hearing impaired people at sentence level has been addressed. A novel method of extracting spatial features to capture hand movements of a signer has been proposed. Frames of a given video of a sign are preprocessed to extract face and hand components of a signer. The local centroids of the extracted components along with the global centroid are exploited to extract spatial features. The concept of interval valued type symbolic data has been explored to capture variations in the same sign made by the different signers at different instances of time. A suitable symbolic similarity measure is studied to establish matching between test and reference signs and a simple nearest neighbour classifier is used to recognize an unknown sign as one among the known signs by specifying a desired level of threshold. An extensive experimentation is conducted on a considerably large database of signs created by us during the course of research work in order to evaluate the performance of the proposed system.

## I. INTRODUCTION

The language used by hearing impaired people for communication is referred to as sign language. The sign language is made up of precise hand movements and facial expressions to convey information, which may not be understood clearly by most of the normal (hearing) people. Hence, there is a huge communication gap between hearing impaired community with normal (hearing) community. Hearing impaired people would need an assistance of another person, who can understand the signs and help them in communicating with normal people by translating the signs into spoken language. It is not feasible to have human translators at all time to assist hearing impaired people in their day-to-day activities. Thus, there is a need of technology support for hearing impaired.

With the advancement in science and technology in general, computer vision in particular, one can think of devising a methodology, which can translate signs into readable text enabling normal people to understand and communicate with hearing impaired people.

In view of this, many researchers in computer vision research community made several attempts to propose a model, which exploits image processing and pattern recognition techniques to recognize signs captured in the form of video. The following section presents some interesting attempts made in this direction from the past two decades.

## II. RELATED WORKS

The research works reported for sign language recognition have addressed the task at finger spelling level [2, 12, 13, 21, 24, 25], at word level [11, 17, 24] and at sentence level [4, 15, 16]. Some of the techniques proposed by the research community, which gained importance due to their performance are Ichetrichef moments [6], Gray level histogram [29], Sensor based glove technique [6,7,10,17], Hidden Morkov Models (HMM) [1], Hu moments and Electromyography (EMG) segmentation [1], Localized contour sequence [10], Size function [17], Transition-movement [5], Moment based size function [8], Convex chain coding and Basic chain code [28], Fourier descriptors [22], Grassman Covariance Matrix (GCM) [31], Fusion of appearance based and 5DT glove based features [19], Sparse Observation (SO) description [27].

From the literature survey, we understand that the models proposed for sign language recognition address the problem either at finger spelling level [20, 24, 26, 27] or at word level [3, 9, 23, 30, 32]. Since signs used by hearing impaired people are very abstract, the sign language recognition based on fingerspelling or word seems to be cumbersome and not effective. With this observation, recently only two attempts were reported to address the problem at sentence level [4, 15, 16]. Therefore, there is scope for many more attempts in this direction.

In view of this, in this research work, we made an attempt to design a model to recognize signs of hearing impaired people at sentence level with some constraints. We have explored the applicability of interval-valued type symbolic data for efficient representation of signs for their recognition.

Rest of the paper is organized as follows: Section 3 presents the proposed method of feature extraction and representation of signs. Experimental results to study the efficacy of the proposed method are presented in section 4, followed by conclusion and future directions in section 5.

## III. THE PROPOSED METHODOLOGY

The proposed sign language recognition system involves four major steps namely (i) Segmentation of hand and face regions from the frames of a given sign video (ii) Extraction of relevant features by establishing spatial relationship among the

Manuscript received March 24, 2016.
Chethana Kumara B M is with the University of Mysore, INDIA, (chethanbm.research@ gmail.com).
Nagendraswamy H S is with University of Mysore, INDIA. (hsnswamy@ compsci.uni-mysore.ac.in).
Lekha Chinmayi R is with University of Mysore, INDIA (lekha.2405@gmail.com).

segmented hand and face components (iii) Compact representation of signs in the knowledgebase (iv) Establishing matching between test and reference signs for recognition. The following subsections present a detailed description about these steps.

### A. Segmentation of Face and Hand Components

In this step, frames are extracted from the given video of signs and are converted into HSV color space. HSV color space being closely associated with human perception provides more information for segmentation compared to RGB color space. Only hue and saturation values from HSV color space are considered to define a threshold value based on local information of a given frame in order to distinguish between skin and non-skin regions. Further, morphological operations such as opening and closing are used iteratively to get more accurately segmented components. Fig. 1 shows the process involved in segmenting hand and face components from the frames of a given sign video. Fig. 2 and Fig. 3 shows the segmented hand and face components for few frames of two different signs as examples.

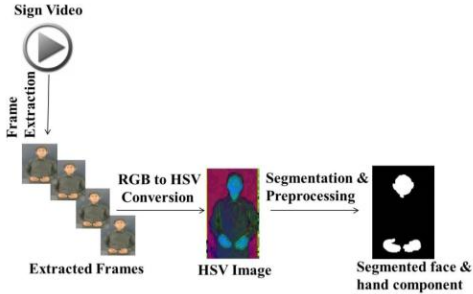

Fig. 1 Segmentation of face and hand components

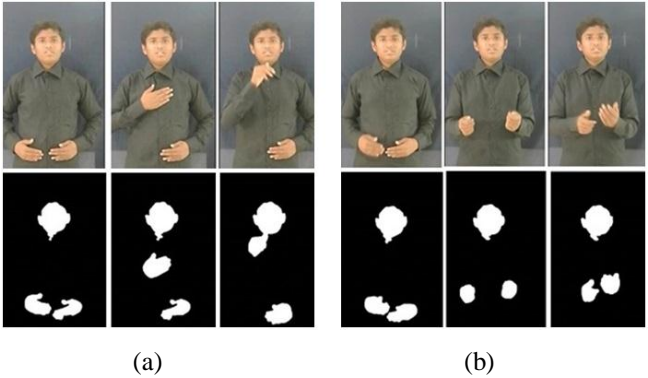

Fig. 2 Segmented face and hand components in various frames of the sentences a) *I WANT COFFEE* and b) *I WANT BUS TICKET*

### B. Feature Extraction

Hand movements of a signer play an important role in making a sign. In most of the signs, the face component is static and one of the hands is moved dynamically over the other. The one, which is more dynamic is called as manual hand and the other one is called as non-manual hand.

It is possible to capture hand movements by establishing spatial relationship among manual hand, non-manual hand and face components in a frame. The spatial relationship captured among the face and hand components in every frame of a sign video defines a signature and is used to characterize the sign. In order to capture spatial relationship, the centroids of segmented face and hand components are computed. Also, the global centroid due to all the three components (face, manual hand, non-manual hand) is computed. Let $C_1$, $C_2$, $C_3$ and $GC$, respectively, denote the centroid of face, manual hand, non-manual hand and global centroid. The distance between global centroid ($GC$) and local centroids ($C_1$, $C_2$, $C_3$) is computed. Also, the distance between ($C_1$ & $C_2$), ($C_1$ & $C_3$) and ($C_2$ & $C_3$) is computed. Let $d_1$, $d_2$ and $d_3$ denote the distance between ($GC$ & $C_1$), ($GC$ & $C_2$) and ($GC$ & $C_3$) respectively. Similarly, let $d_4$, $d_5$ and $d_6$ denote the distance between ($C_1$ & $C_2$), ($C_1$ & $C_3$) and ($C_2$ & $C_3$) respectively. In addition, the angle ($\theta$) made by the line connecting ($C_1$ & $C_2$) and ($C_2$ & $C_3$) is computed. All the distances and angle thus computed are organized in a sequence and considered as feature vector to characterize the spatial relationship among the face and hand components. Fig. 4 illustrates an example of feature extraction from the frame of a given sign video and Fig.5 shows few example instances.

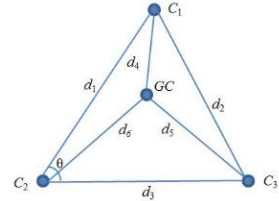

Fig. 4 An instance of feature extraction process

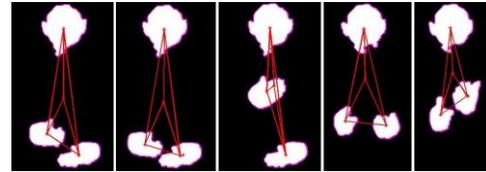

Fig. 5 Few example instances of feature extraction from the frames of a given sign video

While establishing spatial relationship among the components, there are instances where the manual hand overlaps with the face component or with non-manual hand. In such cases, only two components are visible due to overlapping. In order to track the centroid of manual hand when there is a overlapping, we take the mean of the centroid of the manual hand in the previous frame (when there is no overlapping) and the centroid obtained from the overlapped manual hand and face component in the current frame.

Let ($xc_i$, $yc_i$) and ($xc_{i+1}$, $yc_{i+1}$) respectively denote the centroid of manual hand and the centroid of overlapped manual hand and face component in the $i^{th}$ and $(i+1)^{th}$ frame. The centroid for the manual hand in the $(i+1)^{th}$ frame (frame with overlapped manual hand and face component) is computed as

$$xc^1_{i+1} = (xc_i + xc_{i+1})/2$$
$$yc^1_{i+1} = (yc_i + yc_{i+1})/2$$

Since, the face component is almost static, in all the frames, the centroid computed for the face component in the first frame is considered for all the frames.

Similarly, when manual hand overlaps with non-manual hand in the $(i+1)^{th}$ frame, then the individual centroid for manual hand and non-manual hand is computed by taking the mean of their centroid computed for the overlapped manual and non-manual hand in the $(i+1)^{th}$ frame and their respective individual centroids computed in the $(i^{th})$ frame.

Let $(xc_{i1}, yc_{i1})$ and $(xc_{i2}, yc_{i2})$ denote the centroid of the manual and non-manual hand in the ith frame. Let $(xc_{i+1}, yc_{i+1})$ denote the centroid of the overlapped manual and non-manual component in the $(i+1)^{th}$ frame. Then the centroid for manual and non-manual hand in the $(i+1)^{th}$ is computed as

$$xc_{i+1,1} = (xc_{i1} + xc_{i+1})/2$$
$$yc_{i+1,1} = (yc_{i1} + yc_{i+1})/2$$
$$xc_{i+1,2} = (xc_{i2} + xc_{i+1})/2$$
$$yc_{i+1,2} = (yc_{i2} + yc_{i+1})/2$$

Once the centroid for all the three components is known, the features are extracted by establishing spatial relationship among the components as discussed earlier. Fig.6 and Fig.7 shows the illustration of the above process with an example image.

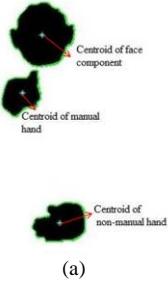
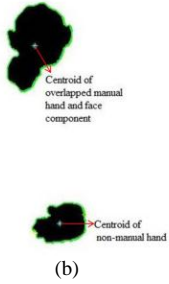
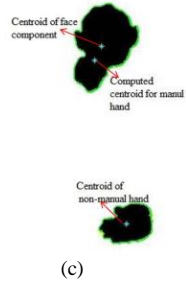

(a)     (b)     (c)

Fig.6(a) Centroids of non-overlapped manual hand, non-manual hand along with face component in the $i^{th}$ frame

Fig. 6 (b) Centroids of manual hand and the overlapped manual hand with face component in the $(i+1)^{th}$ frame

Fig. 6 (c) Computed centroid for manual hand in the $(i+1)^{th}$ frame

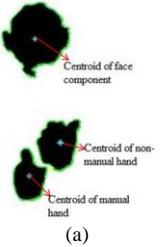
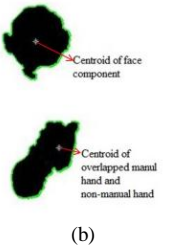
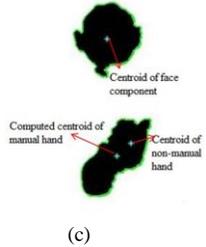

(a)     (b)     (c)

Fig.7(a) Centroids of non-overlapped face and hand components in the $i^{th}$ frame

Fig. 7 (b) Centroids of face component and the overlapped manual and non-manual hand in the $(i+1)^{th}$ frame

Fig. 7 (c) Computed centroid for manual hand and the non manual hand in the $(i+1)^{th}$ frame

The features $d_1, d_2, d_3, d_4, d_5, d_6$ and $\theta$ thus extracted forms a vector $F_1 = \{d_1, d_2, d_3, d_4, d_5, d_6, \theta\}$ to characterize the frames of a sign video.

## C. Sign Representation

Compact representation of sign in the knowledgebase plays an important role in sign recognition system. Issues such as unequal number of frames and intra-class variations must be addressed to provide a robust representation for signs.

The same sign made by the same signer at different instances of time or by different signers may contain unequal number of frames due to the speed at which signs are made and captured. Also, the adjacent frames in the sign video may not differ significantly in terms of their content. Therefore, it is more appropriate to eliminate the frames, which do not differ significantly from previous frames and to select a fixed number of frames for a given sign video, which are sufficient enough to capture sign information irrespective of a signer or instance. In order to address the problem of selection of key frames from sign video, the concept of K-means clustering algorithm is studied. The value of K is empirically chosen after conducting several experimentations. Once the K number of clusters is obtained for each of the signs, each cluster of frames is further processed to select the cluster representative frame. In any cluster, the frame, which possesses maximum similarity with all the frames in the cluster, is considered as a key frame for that cluster. Thus, every sign is characterized by K number of key frames.

In addition to the problem of unequal number of frames for the same sign due to different signers or due to same signer at different instances of time, the instances of signs also possess some variations in terms of their content. In order to capture intra-class variations of a particular sign and to choose multiple representatives for a sign, all the instances of a sign are clustered using hierarchical clustering technique. The idea of inconsistency co-efficient is used during clustering to obtain natural clusters. Inconsistency coefficient values obtained depend on how the samples in a class are clustered at various levels. In order to find the adaptive threshold to cut the dendrogram to obtain natural clusters, we consider the maximum inconsistent coefficient value where all samples are clustered into one cluster and the standard deviation ($\gamma$) of all the non-zero inconsistency coefficient values obtained. The threshold is computed by subtracting the standard deviation ($\gamma$) multiplied by a small constant value ($\delta$) from the maximum inconsistency coefficient values as follows:

Th = max (Inconsistency coefficients) – ($\delta * \gamma$)  (1)

The values for δ are empirically chosen through several experiments.

Once the clusters of various instances of a particular sign are obtained, we consider different percentages of samples (60:40, 50:50, 40:60) within the cluster for the purpose of training and testing. The feature vectors of training samples representing each instance in a cluster are aggregated to form an interval valued type symbolic feature vector, which represent the entire cluster. The min ( ) and max ( ) operations are used to find the minimum and maximum feature values to form interval-valued type feature vector. The process of deriving interval valued type symbolic feature vector for a cluster is described as follows:

Let $S_1, S_2, S_3, ..., S_n$ be the $n$ number of signs considered by the system for study and let $S_i = \{s_1, s_2, s_3, ..., s_m\}$ be the $m$ number of instances of a sign $S_i$ made by the signers at different instances of time.

Let $\{KF_{i1}, KF_{i2}, KF_{i3}, ..., KF_{it}\}$ be the $t$ number of key frames chosen for the video of one of the instances of a sign $S_i$, where $KF_{ij} = \{f_{ij}^1, f_{ij}^2, f_{ij}^3, ..., f_{ij}^l\}$ be the feature vector representing $j^{th}$ key frame of one of the instances of a sign $S_i$, and $l$ is the number of features.

Let $c$ be the number of clusters obtained from $m$ instances of a sign $S_i$. If a particular cluster say $p$ among $c$ number of clusters contain $q$ number of instances, then the features describing the $j^{th}$ key frame of all the $q$ number of instances are aggregated to form an interval type symbolic data as described below.

Let

$$KF_{ij}^{(1)} = \{KF_{ij}^{(1)1}, KF_{ij}^{(1)2}, KF_{ij}^{(1)3}, ..., KF_{ij}^{(1)l}\}$$
$$KF_{ij}^{(2)} = \{KF_{ij}^{(2)1}, KF_{ij}^{(2)2}, KF_{ij}^{(2)3}, ..., KF_{ij}^{(2)l}\}$$
$$KF_{ij}^{(3)} = \{KF_{ij}^{(3)1}, KF_{ij}^{(3)2}, KF_{ij}^{(3)3}, ..., KF_{ij}^{(3)l}\}$$
$$\vdots$$
$$KF_{ij}^{(q)} = \{KF_{ij}^{(q)1}, KF_{ij}^{(q)2}, KF_{ij}^{(q)3}, ..., KF_{ij}^{(q)l}\}$$

be the feature vectors representing the $j^{th}$ key frame of the $1^{st}$, $2^{nd}$, $3^{rd}$, ..., $q^{th}$ instances of a cluster $p$, respectively. Then,

$$f_{ij}^{1-} = Min\{f_{ij}^{(1)1}, f_{ij}^{(2)1}, f_{ij}^{(3)1}, ..., f_{ij}^{(q)1}\}$$
$$f_{ij}^{1+} = Max\{f_{ij}^{(1)1}, f_{ij}^{(2)1}, f_{ij}^{(3)1}, ..., f_{ij}^{(q)1}\}$$

Similarly, we compute

$$f_{ij}^{2-}, f_{ij}^{2+}, f_{ij}^{3-}, f_{ij}^{3+}, ....., f_{ij}^{l-}, f_{ij}^{l+}$$

Thus, the aggregated $j^{th}$ key frame of reference feature vector representing the $p^{th}$ cluster of a sign $S_i$ is given by

$$RF_{ij}^p = \{[f_{ij}^{(p)1-}, f_{ij}^{(p)1+}], [f_{ij}^{(p)2-}, f_{ij}^{(p)2+}], [f_{ij}^{(p)3-}, f_{ij}^{(p)3+}], ..., [f_{ij}^{(p)l-}, f_{ij}^{(p)l+}]\}$$

### D. Matching and Recognition

In order to recognize a given test sign made by the signer, the video sequence of a test sign is processed to obtain frames, and the features are extracted from each frame as discussed in the previous section. The extracted features are organized in a sequence to represent the test sign. Since the test sign involves only one instance, the test sign is represented in the form of a crisp feature vector.

The task of recognition is accomplished by comparing the test sign feature vector with all the reference sign feature vectors stored in the knowledgebase. A similarity value is computed through this process and the reference sign, which possess maximum similarity value with the test sign, is considered and the text associated with this sign is displayed. The similarity measure proposed in [14] is used for the purpose of comparing reference sign feature vector with the test sign feature vector as follows

Let $TF_j = \{f_j^1, f_j^2, f_j^3, ..., f_j^l\}$ be the crisp feature vector and $RF_j = \{[f_j^{1-}, f_j^{1+}], [f_j^{2-}, f_j^{2+}], [f_j^{3-}, f_j^{3+}], ..., [f_j^{l-}, f_j^{l+}]\}$ be the interval valued type symbolic feature vector representing the $j^{th}$ key frame of a test and a reference sign respectively.

Similarity between the test and reference sign with respect to $j^{th}$ key frame is computed as

$$SIM(RF_j, TF_j) = \frac{1}{l} \sum_{d=1}^{l} \left\{ \max\left[ \frac{1}{1+abs(f_j^{d-} - f_j^d)}, \frac{1}{1+abs(f_j^{d+} - f_j^d)} \right] \begin{array}{l} if(f_j^{d-} \leq f_j^d \leq f_j^{d+}) \\ otherwise \end{array} \right\} \quad (2)$$

The total similarity between the test and reference sign due to all the frames is computed as

$$SIM(RF, TF) = \sum_{j=1}^{L} SIM(RF_j, TF_j) \quad (3)$$

where $L$ is the number frames used to represent the test and reference sign, which is 40. The nearest neighbor classification technique is used to recognize the given test sign as one among the known sign in the sign knowledgebase.

## IV. EXPERIMENTATION

In order to demonstrate the efficacy of the proposed method, we have conducted experiments on UoM-ISL sign language dataset. The dataset describes the sentences used by hearing impaired people in their day to day life. We have considered the videos of signs, which are signed by the hearing impaired students of different schools of Mysore zone. The data set contains 1040 (17.3 hours) sign videos of 26 different signs expressed by four different students with ten instances.

The values for $\delta$ in Eqn. (1) can be in the range 0.1 to 1.0. However, the number of clusters is significantly changed for only some values of $\delta$. In our experiments, we found that the number of clusters obtained for $\delta = (1, 0.5, 0.1)$ are significantly different and hence, we chose these three values for three different experiments on the dataset of signs considered.

Several experiments are conducted for different percentages of training and testing (60:40, 50:50 and 40:60). We have also repeated the experiment with random sets of training and testing samples chosen from each cluster. In each experiment, performance of the system is measured in terms of F-measure.

Table I gives the average F-measure of all 50 random trails for each class for different percentages of training and testing samples for one of the three different numbers of clusters (357 representatives). Table II gives the overall average recognition performance of the proposed method in terms of F-measure for different percentages of training and testing samples for the entire database and for three different numbers of clusters.

TABLE I
AVERAGE RECOGNITION RATE OF THE PROPOSED METHOD FOR 357 REPRESENTATIVES

| Class Index | Training : Testing | | | Class Index | Training : Testing | | |
|---|---|---|---|---|---|---|---|
| | 40:60 | 50:50 | 60:40 | | 40:60 | 50:50 | 60:40 |
| 1 | 0.85 | 0.79 | 0.80 | 14 | 0.84 | 0.90 | 0.88 |
| 2 | 0.74 | 0.77 | 0.74 | 15 | 0.88 | 0.90 | 0.94 |
| 3 | 0.73 | 0.81 | 0.73 | 16 | 0.96 | 0.97 | 0.97 |
| 4 | 0.77 | 0.64 | 0.67 | 17 | 0.93 | 0.97 | 0.95 |
| 5 | 0.71 | 0.73 | 0.67 | 18 | 0.83 | 0.92 | 0.85 |
| 6 | 0.77 | 0.80 | 0.67 | 19 | 0.85 | 0.83 | 0.92 |
| 7 | 0.83 | 0.77 | 0.87 | 20 | 0.96 | 0.96 | 0.86 |
| 8 | 0.79 | 0.83 | 0.82 | 21 | 0.94 | 0.91 | 0.97 |
| 9 | 0.92 | 1.00 | 0.96 | 22 | 0.80 | 0.92 | 0.83 |
| 10 | 0.90 | 0.89 | 0.93 | 23 | 0.87 | 0.87 | 0.90 |
| 11 | 0.85 | 0.90 | 0.88 | 24 | 0.77 | 0.84 | 0.77 |
| 12 | 0.81 | 0.85 | 0.70 | 25 | 0.80 | 0.81 | 0.79 |
| 13 | 0.93 | 0.88 | 0.82 | 26 | 0.73 | 0.90 | 0.85 |

TABLE II
OVERALL AVERAGE RECOGNITION PERFORMANCE OF THE PROPOSED METHOD

| Sign Language Data Set | Ratio of Training and Testing | No. of representatives | Overall Average Recognition Rate |
|---|---|---|---|
| UoM-ISL Data | 60:40 | 286 | 82.13±0.19 |
| | | 312 | 83.70±0.61 |
| | | 357 | **85.09±0.78** |
| | 50:50 | 256 | 80.99±0.98 |
| | | 312 | 81.35±0.59 |
| | | 357 | **82.86±0.83** |
| | 40:60 | 256 | 76.11±1.23 |
| | | 312 | 77.92±0.68 |
| | | 357 | **79.07±1.11** |

Fig. 8 to Fig. 10 shows the confusion matrices obtained for 60:40, 50:50 and 40:60 percentages of training and testing samples respectively for one of the random runs. Performance of the proposed system with crisp representation has also been studied interms of leave-one-out and 10 fold Cross validation techniques and the results are presented in Table III. In leave-one-out classification technique, out of 1040 signs (videos), 1014 signs are used as reference signs in the knowledgebase and 26 signs are used for testing the classification performance of the system. Similarly, in 10-fold cross validation technique, 780 signs out of 1040 are used as reference signs and 260 are used for testing. In both the validation techniques, testing samples were chosen randomly for different run and the overall average classification accuracy in terms of F-measure for the entire database is reported.

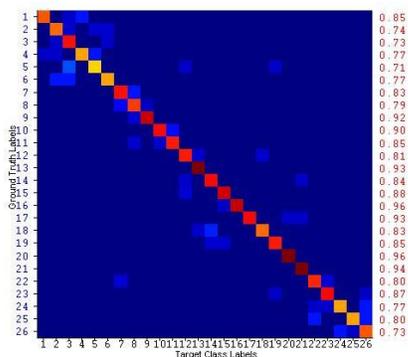

Fig. 8 Confusion matrix for (50:50) training and testing samples

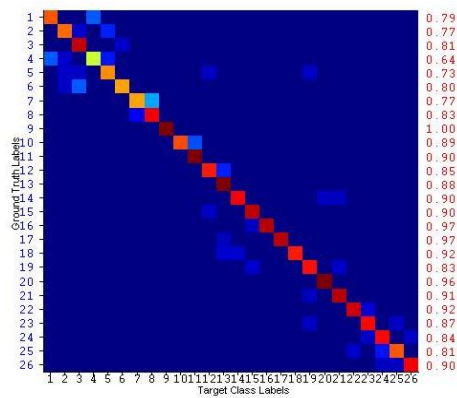

Fig. 9 Confusion matrix for (40:60) training and testing samples

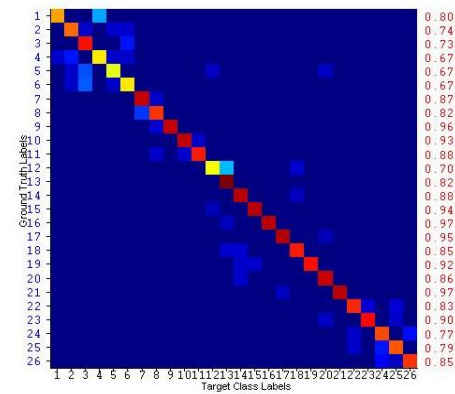

Fig. 10 Confusion matrix for (60:40) training and testing samples

In the proposed system with symbolic representation technique, out of 1040 signs, 357 signs selected through clustering process are used as reference signs in the knowledgebase and the remaining 416 samples are used for testing the system (for 60:40 training and testing).

When compared to other validation techniques, the proposed technique uses less number of reference signs in the knowledgebase. Thus, we claim that the proposed method is efficient in terms of storage and robust in terms of representation as it effectively captures the intra-class variations in the form of interval type data.

TABLE III
RESULTS OBTAINED BY THE PROPOSED METHODOLOGY FOR DIFFERENT VALIDATION TECHNIQUES

| Different Validation techniques | Number of reference signs stored in knowledgebase | Recognition rate (Average F-measure) |
|---|---|---|
| Leave-one-out | 1014 | 76.25 ± 1.75 |
| 10 fold Cross validation | 780 | 75.16 ± 1.20 |

Experiments were also conducted to study the performance of the proposed system for signers' independent sign recognition. Out of four signers, signs made by three signers are used to train the system and the signs made by the other signer are used for testing. Thus, out of 1040 signs, 780 signs are used for training and 260 signs for testing. The same experiment is repeated four times, where at each time signs made by one of the signers are used for testing while the signs made by the other three signers are used to train the system.

Thus, a leave-one-out classification technique has been followed in this experiment. Performance of the system has been measured in terms of F-measure for the entire database and the results are presented in Table IV.

TABLE IV
PERFORMANCE OF THE PROPOSED METHOD FOR DIFFERENT PERCENTAGES OF TRAINING AND RESTING SAMPLES FOR SIGNER INDEPENDENT SIGN RECOGNITION

| Sign Language Data Set | Average recognition rate | | |
|---|---|---|---|
| | Training | Testing | Average F-measure |
| UoM-ISL Data | 2, 3, 4 | 1 | 59.91±0.83 |
| | 1, 3, 4 | 2 | 58.51±0.43 |
| | 1, 2, 4 | 3 | 58.90±0.88 |
| | 1, 2, 3 | 4 | 57.97±0.67 |
| Overall average recognition rate | | | 58.82 |

## V. CONCLUSION

In this work, we made an attempt to recognize the signs of hearing impaired people at sentence level. Signs are captured in the form of video and each frame in the video is processed to characterize the hand movements of a signer in terms of spatial features. *K*-means clustering technique has been exploited to address unequal number of frames in a video and every sign is characterized in terms of a fixed number of key frames. An interval valued type symbolic data has been explored to capture intra-class variations in a sign due to many practical considerations. Suitability of symbolic similarity measure and first nearest neighbour classification technique has been studied for sign matching and recognition. Performance of the proposed methodology has been corroborated by conducting extensive experiments on considerably large sign corpus. Superiority of the proposed validation technique has been established by comparing with the other standard validation techniques such as 10 fold Cross validation and leave-one-out validation. Performance of the proposed system for signer-independent sign recognition has also been studied. An average F-measure is used as a metric for comparison in all the experiments.

After analyzing the results obtained by various experiments, we conclude that the proposed approach is effective in capturing signers hand movements to characterize the sign in a simple way. However, the method may not be able to distinguish signs if they have same hand movements but differ in their shape structure to convey different message. Hence, performance of the system may drop in such cases. Also scalability of the proposed approach needs to be studied considering a large database of sign with complex sentences.

## ACKNOWLEDGEMENT

We would like to thank the students and the teaching staff of Sai Ranga Residential Boy's School for Hearing Impaired, Mysore, and N K Ganpaiah Rotary School for physically challenged, Sakaleshpura, Hassan, Karnataka, INDIA, for their immense support in the process of Sign language data set creation.

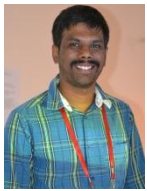
**Chethana Kumara B M** obtained his MCA from University of Mysore, Mysuru, Karnataka, India in 2010. He is currently working towards his Ph.D degree in the area of computer vision at University of Mysore. His focused area of research is Sign Language Recognition, Gesture recognition and Texture analysis.

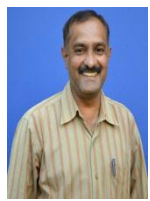
**Nagendraswamy H S** obtained his M.Sc and Ph.D degrees from University of Mysore, India in 1994 and 2007 respectively. He is currently working as Associate Professor in the Department of Studies in Computer Science, University of Mysore, Manasagangothri, Mysore, Karnataka, India. His focused areas of research include Shape analysis, Texture analysis, Sign Language Recognition, Precision agriculture, Symbolic data analysis, Fuzzy theory, Biometrics and Video analysis.

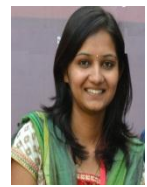
**Lekha Chinmayi R** obtained her MCA from University of Mysore, Mysuru,Karnataka, India in 2013. She is currently working as project associate in the Department of Studies in Computer Science, University of Mysore. Her focused area of research is Sign Language Recognition